\documentclass[letterpaper]{article} 
\usepackage{aaai2027}  
\usepackage[hyphens]{url}  
\usepackage{graphicx} 
\usepackage{natbib}  
\usepackage{caption} 
\usepackage{algorithm}

\usepackage{newfloat}
\usepackage{listings}
\DeclareCaptionStyle{ruled}{labelfont=normalfont,labelsep=colon,strut=off} 
\floatstyle{ruled}
\newfloat{listing}{tb}{lst}{}
\floatname{listing}{Listing}

\usepackage{graphicx}%
\usepackage{multirow}%
\usepackage{amsmath,amssymb,amsfonts}%
\usepackage{mathrsfs}%
\usepackage[title]{appendix}%
\usepackage{xcolor}%
\usepackage{textcomp}%
\usepackage{manyfoot}%
\usepackage{booktabs}%
\usepackage{algorithm}%
\usepackage{algorithmicx}%
\usepackage{algpseudocode}%
\usepackage{listings}%
\usepackage{booktabs}
\usepackage{bm} 
\def\mc{\mathcal}

\usepackage{caption}

\usepackage{enumitem}
\usepackage{graphicx} 
\usepackage{appendix}
\usepackage{arydshln}  
\usepackage{booktabs}

\title{DeCLIP: Decoupled Prompting for Multi-Label \\ Class-Incremental Learning with CLIP}
\author{
    Kaile Du\textsuperscript{\rm 1}, Zihan Ye\textsuperscript{\rm 2}, Junzhou Xie\textsuperscript{\rm 1}, Yixi Shen\textsuperscript{\rm 3}, Yuyang Li\textsuperscript{\rm 1}, Fuyuan Hu\textsuperscript{\rm 3}, Ling Shao\textsuperscript{\rm 2},\\ Guangcan Liu\textsuperscript{\rm 1}\corresponding, Joost van de Weijer\textsuperscript{\rm 4}, Fan Lyu\textsuperscript{\rm 4}\corresponding
}
\affiliations{
    \textsuperscript{\rm 1}School of Automation, Southeast University, China

    \textsuperscript{\rm 2}University of the Chinese Academy of Sciences, China

    \textsuperscript{\rm 3}Suzhou University of Science and Technology, China

    \textsuperscript{\rm 4}Computer Vision Center, Spain

}

\begin{document}

\maketitle

\begin{abstract}
Multi-label class-incremental learning (MLCIL) continuously expands the label space while recognizing multiple co-occurring categories, making catastrophic forgetting a central challenge. Recent class-incremental learning methods have increasingly adopted CLIP as their backbone. However, we find that applying CLIP to MLCIL exhibits two critical issues: entanglement of class-specific cues in shared visual representations and high false-positive rates (FPR) under task-level partial labeling. We propose DeCLIP, a replay-free and parameter-efficient framework for CLIP-based MLCIL. DeCLIP uses Decoupled Prompting to learn class-specific positive and negative prompts in both visual and textual modalities, enabling class-conditioned vision-language matching and reducing representation entanglement. Only new-category prompts are optimized; previous prompts remain unchanged, preserving prior knowledge and mitigating catastrophic forgetting without replay. DeCLIP further incorporates Adaptive Similarity Tempering, an inference-time strategy that adapts similarity-tempering strength to the incremental configuration, suppressing false positives without specific tuning. Experiments on MS-COCO, PASCAL VOC, and the real-world NUS-WIDEseq benchmark demonstrate consistent improvements over prior methods with a few trainable parameters.
\end{abstract}

\begin{figure}[t]
    \centering
    \includegraphics[width=\linewidth]{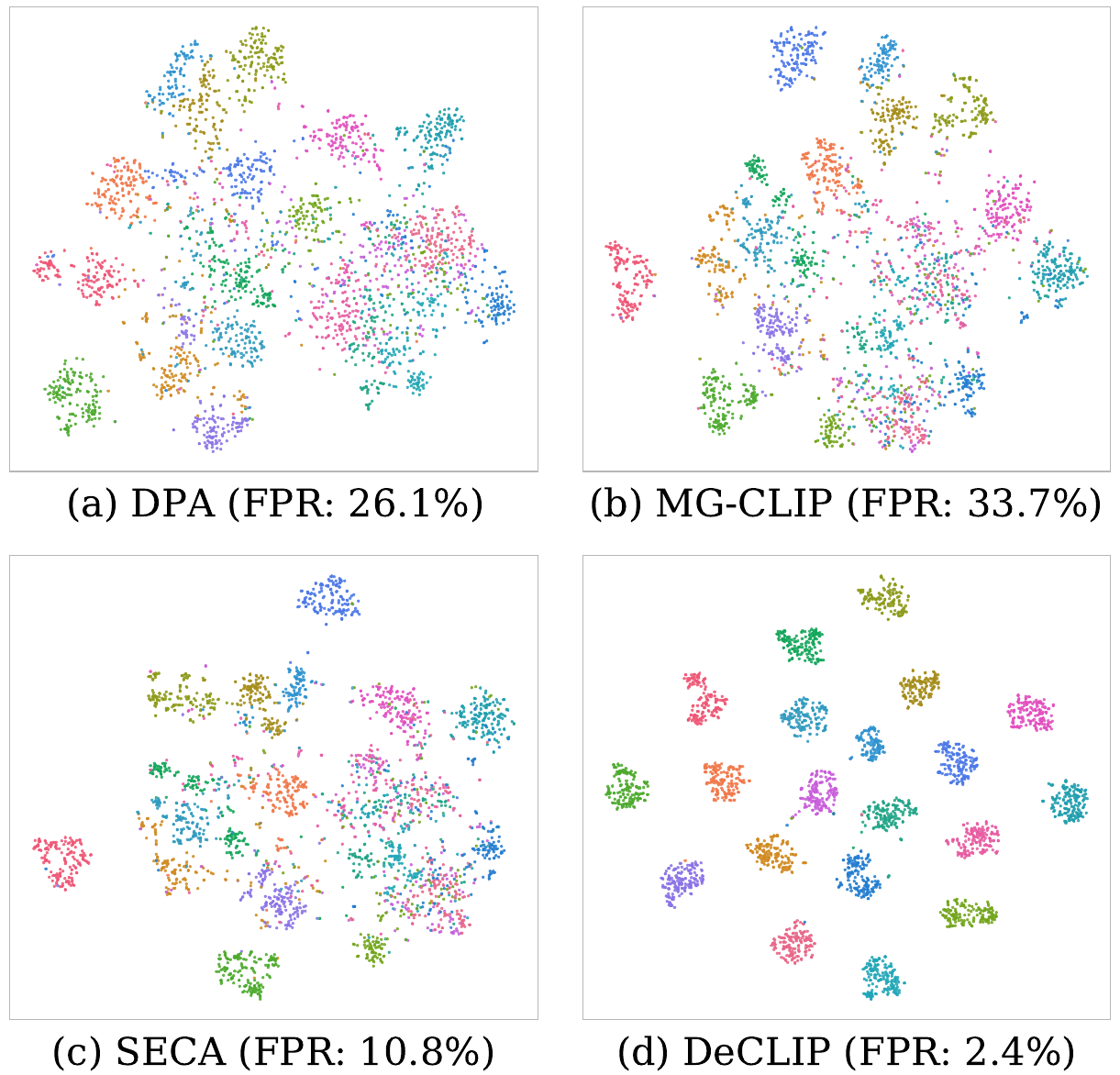}
 \caption{
    t-SNE visualization of category-wise prediction features on VOC B4-C2 after the final incremental task. Compared with DPA, MG-CLIP, and SECA, DeCLIP forms more separable category clusters and achieves a markedly lower false-positive rate (FPR).  DP alleviates representation entanglement, while AST effectively suppresses FPR.
    }
    \label{fig:figure1}
\end{figure}

\section{Introduction}

Class-incremental learning (CIL) aims to acquire new categories sequentially
while retaining the ability to recognize those learned previously. Its central
challenge is catastrophic forgetting, where learning from new data degrades
performance on earlier categories~\cite{kirkpatrick2017overcoming}.
Multi-label class-incremental learning (MLCIL) extends this setting to images
containing multiple co-occurring categories~\cite{dong2023knowledge}. It is further complicated by task-level partial labeling
~\cite{du2024confidence,du2025rebalancing}: only the categories of the current
task are annotated, while annotations for categories from previous and future
tasks are unavailable. Consequently, MLCIL must continually expand its label
space under limited supervision while preserving knowledge of previously
learned categories.

Recent CIL methods increasingly adopt CLIP~\cite{radford2021learning} through
prompts, adapters, or lightweight classifiers~\cite{jha2024clap4clip,huang2024class,yu2024boosting,huang2025mind,he2026harnessing}. Most of these methods are designed for single-label CIL, where an image is
associated with one target category. When adapting CLIP to MLCIL, we observe two additional difficulties. First,
CLIP typically encodes an image into a shared set of visual features and
compares these features with class-specific textual representations. In a multi-label image, using the class-shared visual features for all
category-wise comparisons can entangle class-specific cues. As shown in
Figure~\ref{fig:figure1}, DPA~\cite{zhao2024dynamic},
MG-CLIP~\cite{huang2025mind}, and SECA~\cite{he2026harnessing} exhibit
substantial category overlap in the representations used for category-wise
prediction, revealing pronounced \textbf{representation entanglement}.
Such entanglement makes the evidence associated with different co-occurring
categories difficult to distinguish. Moreover, continual adaptation to
incoming categories may weaken evidence useful for previously learned
categories, further aggravating catastrophic forgetting.
Second, task-level partial labeling leaves many negative
classes unsupervised, causing the model to assign excessive
confidence to absent categories and produce \textbf{false-positive
predictions}~\cite{du2025rebalancing}. We further observe that this phenomenon is  pronounced in CLIP-based
methods. As shown in Figure~\ref{fig:figure1}, the three  methods
produce FPRs ranging from 10.8\% to 33.7\%.

To address these issues, we propose \textbf{DeCLIP}, a replay-free and
parameter-efficient framework for CLIP-based MLCIL. DeCLIP introduces
\textbf{Decoupled Prompting} (DP), which learns class-specific positive and
negative prompts in both visual and textual modalities. To keep this
class-specific adaptation parameter-efficient, DeCLIP uses compact prompts and
applies visual prompting only in late encoder blocks, with the same prompt shared across these blocks. For each category,
visual prompts guide the frozen image encoder to extract a
class-conditioned representation, while textual prompts describe the
presence and absence semantics. Matching the corresponding
visual and textual representations enables DeCLIP to focus on
category-relevant evidence, thereby reducing representation entanglement among
co-occurring categories. Since only current labels are available, DeCLIP updates current prompts while retaining previous prompts, mitigating forgetting without replay.
DeCLIP further incorporates Adaptive Similarity Tempering (AST),
an inference-time strategy that adaptively adjusts the strength of
positive--negative similarity tempering according to the incremental
configuration. AST applies weaker tempering to configurations
requiring less FPR control and stronger tempering to long-sequence
configurations with greater false-positive bias, without dataset-
or scenario-specific hyperparameter tuning. As shown in
Figure~\ref{fig:figure1}, DeCLIP produces more compact and separable
class-conditioned representations, while AST reduces the FPR to 2.4\%.

Extensive experiments on MS-COCO, PASCAL VOC, and the real-world NUS-WIDEseq
 demonstrate the effectiveness of DeCLIP. Our
contributions are summarized as follows:

\begin{enumerate}
    \item We propose DeCLIP for replay-free MLCIL. DP reduces representation entanglement using class-specific prompts in both modalities. Its class-wise optimization updates only current prompts and retains previous ones, mitigating forgetting without replay.

    \item We incorporate AST into DeCLIP to adapt false-positive
suppression strength to the incremental configuration, without dataset-
or scenario-specific tuning.

    \item Extensive experiments on MS-COCO, PASCAL VOC, and the real-world
    NUS-WIDEseq benchmark demonstrate that DeCLIP achieves consistent
    improvements with only a small number of trainable parameters.
\end{enumerate}

\section{Related Work}

\noindent\textbf{Single-label class-incremental learning (SLCIL).}
SLCIL aims to learn new categories while preserving knowledge. Regularization-based methods mitigate forgetting by constraining
model updates~\cite{kirkpatrick2017overcoming,li2017learning},
whereas replay-based methods retain a memory of previous samples
\cite{rolnick2019experience}. Prompt-based
methods adapt pre-trained models with lightweight learnable
parameters. L2P~\cite{wang2022learning} and DualPrompt~\cite{dualprompt2022wang}
retrieve prompts from a learned pool for different inputs.
Vision--language models have also become increasingly popular in SLCIL.
CLAP~\cite{jha2024clap4clip} adopts probabilistic modeling, MOE4CL
\cite{yu2024boosting} uses a mixture-of-experts architecture with routing, and
RAPF~\cite{huang2024class} introduces a lightweight visual adaptation module.
MG-CLIP~\cite{huang2025mind} addresses the modality gap through a compensation
mechanism, while SECA~\cite{he2026harnessing} exploits textual semantic priors
and visual prototype refinement for knowledge transfer.

\noindent\textbf{Multi-label class-incremental learning (MLCIL).}
MLCIL additionally requires recognition of multiple categories
under task-level partial labeling. Replay-based methods include PRS~\cite{kim2020imbalanced} and
CUTER~\cite{wang2025cut}, while KAR~\cite{CAO2026133167} introduces a
knowledge-aware replay framework. Among replay-free methods,
KRT~\cite{dong2023knowledge} transfers previous knowledge, and
CSC~\cite{du2024confidence} recalibrates label dependencies. RebLL~\cite{du2025rebalancing} handles positive--negative
imbalance through asymmetric loss. L3A \cite{zhang2025l3a} proposes label-augmented analytic adaptation for MLCIL, but adopts a different evaluation criterion. N-CGCN~\cite{du2026negative} further explores negative-aware distillation. 
Recent work has begun to adapt prompt-based methods to MLCIL.
MULTI-LANE~\cite{de2024less} uses patch selection and task prompts.
DPA~\cite{zhao2024dynamic} extends DualCoOp~\cite{sun2022dualcoop} to the MLCIL setting, and incorporates
confidence-based exemplar replay. We also use its zero-buffer version in
our replay-free comparison.


    \begin{figure*}[t]
        \centering
        \includegraphics[width=\linewidth]{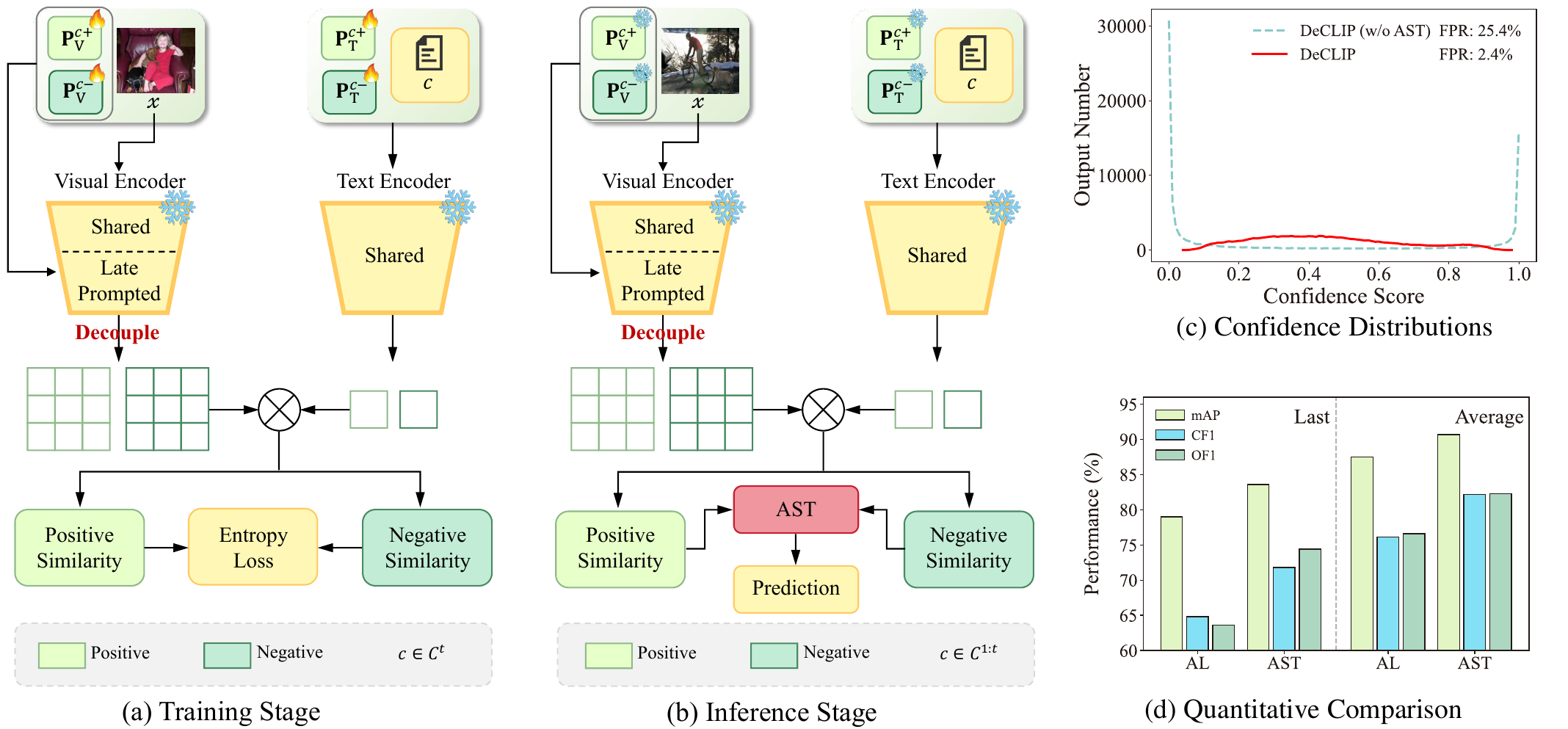}
                \caption{(a) DeCLIP training on task $t$. For each class $c \in \mc{C}^{t}$, positive and negative prompts guide the frozen encoders to decouple class-specific components. The image first passes through a shared frozen prefix,
and class-specific visual prompts are injected only into the last $n$ blocks; the text
encoder is shared. (b) DeCLIP inference after task $t$. All learned prompt pairs for classes
$c\in\mathcal{C}^{1:t}$ are preserved. AST adjusts the tempering strength of  
similarities according to the incremental configuration. (c) Confidence distributions with and without AST on VOC B4-C2, where AST reduces the FPR from 25.4\% to 2.4\%. (d)  Comparison of AL and AST on the last and average performance in the
long-sequence VOC B4-C2 setting.
}
        \label{fig:figure2}
    \end{figure*}
\section{Method}

\subsection{Preliminary}
MLCIL is defined over a sequence of $T$ tasks. 
Each task $t \in \{1,\dots,T\}$ is associated with a training set $\mc{D}^{t}_\text{trn}$ and a testing set $\mc{D}^{t}_\text{tst}$, with its own label space $\mc{C}^t$. 
The cumulative label space up to task $t$ is denoted as $\mc{C}^{1:t} = \bigcup_{i=1}^{t} \mc{C}^i$, where the label sets are disjoint, $\mc{C}^{i} \cap \mc{C}^{j} = \varnothing$ for $i \neq j$. 
Under the task-level partial labeling setting, only the current label set $\mc{Y}^t_\text{trn} = \mc{C}^t$ is available during training. 
At inference step $t$, the model is required to predict over the expanded label space $\mc{Y}^t_\text{tst} = \mc{C}^{1:t}$, ensuring that after completing task $t$ it can recognize all categories encountered so far.

\subsection{DeCLIP Overview}

Figure~\ref{fig:figure2}(a) and (b) present the overall framework of DeCLIP. DeCLIP is
built on a frozen CLIP model and consists of two  components:
DP and AST. DP
learns class-specific positive and negative prompts in both visual and textual
modalities to reduce representation entanglement among co-occurring categories.
At each task, only current-category prompts are optimized; previous prompts remain unchanged and are reused at inference, preserving prior knowledge without replay.
AST is applied at inference to adapt the tempering strength of the
 positive--negative similarities and suppress
false positives. To remain parameter-efficient, we further adopt compact, layer-shared visual prompting.

\subsection{Decoupled Prompting}

CLIP typically encodes an image into a shared set of visual features and
compares these features with the class-specific  textual representations. When applied to MLCIL, however, the same image may contain several
co-occurring categories, and using the same visual features for all
vision--language comparisons can entangle their class-specific cues. \textbf{Decoupled prompting} (DP) 
addresses this representation entanglement by assigning each class its own
visual and textual prompt pair. These prompts condition the  CLIP
encoders on a specific class, enabling each vision--language comparison to
operate on a class-conditioned representation rather than a common visual
representation shared by all class
decisions.

Specifically, for each category $c$, we construct a class-specific prompt set
$\mathcal{P}^{c}=\{\mathbf{P}^{c}_{\mathrm{T}},
\mathbf{P}^{c}_{\mathrm{V}}\}$, where
$\mathbf{P}^{c}_{\mathrm{T}}=\{\mathbf{P}^{c+}_{\mathrm{T}},
\mathbf{P}^{c-}_{\mathrm{T}}\}$ denotes the positive and negative text
prompts, and
$\mathbf{P}^{c}_{\mathrm{V}}=\{\mathbf{P}^{c+}_{\mathrm{V}},
\mathbf{P}^{c-}_{\mathrm{V}}\}$ denotes the corresponding visual prompts.
The positive prompts model the presence of category $c$, whereas the negative
prompts model its absence. The visual and textual prompts are jointly used to
produce a positive--negative comparison. As shown in Figure~\ref{fig:figure2}(a), 
for an input image $x$, the text encoder combines a class name with its
positive or negative text prompt to obtain a text representation. 
The visual encoder receives the image together with the corresponding visual
prompt and produces a class-conditioned visual representation. We compute
the positive and negative vision--language similarities as
\begin{equation}
\label{eq:t_p}
\begin{aligned}
s^{c+} &=
\cos\!\left(
\mathrm{E}_{\mathrm{T}}(\mathbf{P}_{\mathrm{T}}^{c+},c),
\mathrm{E}_{\mathrm{V}}(\mathbf{P}_{\mathrm{V}}^{c+},x)
\right),\\[3pt]
s^{c-} &=
\cos\!\left(
\mathrm{E}_{\mathrm{T}}(\mathbf{P}_{\mathrm{T}}^{c-},c),
\mathrm{E}_{\mathrm{V}}(\mathbf{P}_{\mathrm{V}}^{c-},x)
\right).
\end{aligned}
\end{equation}

Here, $\mathrm{E}_{\mathrm{T}}(\cdot)\in\mathbb{R}^{d}$ produces a text
embedding, whereas
$\mathrm{E}_{\mathrm{V}}(\cdot)\in\mathbb{R}^{L\times d}$ produces $L$ visual
tokens for image $x$, where $d$ is the shared embedding dimension. Following
common practice~\cite{sun2022dualcoop,hu2023dualcoop++,zhao2024dynamic}, we
compare the text embedding with each visual token, aggregate the token-wise
cosine similarities, and apply the CLIP logit scale to obtain the scores $s^{c+}$ and $s^{c-}$.
$\cos(\cdot,\cdot)$ denotes the token-wise aggregated cosine
similarity.
Consequently, different category decisions rely on their respective
class-conditioned visual evidence, thereby reducing representation
entanglement among co-occurring categories.

The two scores are normalized with a binary softmax:
\begin{equation}
\label{eq:softmax_pos}
\hat{y}_{c+}^{t} =
\frac{\exp\!\left(s^{c+}/\tau(t)\right)}
{\exp\!\left(s^{c+}/\tau(t)\right)+
 \exp\!\left(s^{c-}/\tau(t)\right)}.
\end{equation}
Here, $\hat{y}_{c+}^{t}$ is the confidence that category $c$ is present in
image $x$, and
$\hat{y}_{c-}^{t}=1-\hat{y}_{c+}^{t}$ is the corresponding negative
confidence. We set $\tau(t)=1$ during training. At inference, AST adjusts this
temperature to rescale the confidence, as described in the next subsection.

Under partial labeling, DeCLIP optimizes only prompts for $c \in \mc{C}^{t}$. Previous prompts receive no gradients and remain unchanged, but are reused over $\mc{C}^{1:t}$ at inference. This naturally mitigates forgetting without replay or an auxiliary retention loss.
As shown in Figure~\ref{fig:figure2}(b), inference uses all
prompt sets learned so far, allowing DeCLIP to recognize previously encountered
and newly introduced categories without replay.

\subsection{Adaptive Similarity Tempering}
DP extracts category-related evidence but does not explicitly control the
false-positive bias caused by task-level partial labeling. At each stage,
only current categories are supervised, while labels for old categories are
unavailable. Although some unobserved labels may be positive, most
unobserved category--image pairs are negative; the resulting lack of negative supervision can cause the model to assign excessive confidence to absent categories~\cite{du2025rebalancing}. For example, DeCLIP without AST
reaches an FPR of 25.4\% on VOC B4-C2
(Figure~\ref{fig:figure2}(c)).

Building on prior MLCIL findings
\cite{du2024confidence,du2025rebalancing}, we find that the need for FPR
suppression varies across incremental configurations. Long-sequence MLCIL settings supervise fewer categories per stage relative to the
accumulated label space and tend to exhibit stronger false-positive bias.
Conversely, configurations with a large base task or larger increments may
require little suppression, where unnecessary tempering can weaken valid
positive predictions.


We therefore introduce \textbf{Adaptive Similarity Tempering
(AST)}, an inference-time strategy that adaptively adjusts the
strength of positive--negative similarity tempering according to
the incremental configuration. Configurations requiring less FPR
control receive weaker tempering, whereas long-sequence
configurations with greater false-positive bias receive stronger
tempering. AST does not modify the learned prompts or require dataset- or
scenario-specific tuning.

AST builds on the binary softmax in Eq.~\eqref{eq:softmax_pos}. For each
class, it applies an adaptive temperature to the positive--negative
similarities $(s^{c+},s^{c-})$:
\begin{equation}
\label{eq:AST}
\tau(t)=
\begin{cases}
1, & t=1,\\[4pt]
\max\!\left(
\lambda\,\dfrac{t}{|\mathcal{C}^{1:t}|},\,1
\right), & \text{otherwise},
\end{cases}
\end{equation}
where $|\mathcal{C}^{1:t}|$ is the seen classes number.  $\lambda$
is a global constant shared across datasets and scenarios and requires no specific tuning, following
\cite{cheng2024towards}. The ratio
$t/|\mathcal{C}^{1:t}|$ characterizes incremental configuration and is larger
for sequences introducing fewer classes per task, where false positives are
typically more severe.

When $\lambda t/|\mathcal{C}^{1:t}|\leq1$, AST sets
$\tau(t)=1$ and recovers the original predictions. This weakest
tempering case occurs throughout COCO B40-C10. In long-sequence
settings such as VOC B4-C2, $\tau(t)>1$ after the base task, and a
larger temperature applies stronger tempering to overconfident
positive--negative comparisons.
As shown in Figure~\ref{fig:figure2}(c), AST reduces the heavy tail
of overconfident predictions and lowers the FPR from 25.4\% to
2.4\% on VOC B4-C2. Unlike AL~\cite{du2025rebalancing}, which modifies the training
objective and may weaken positive-label learning, AST instead tempers
positive--negative similarities at inference without interfering with
prompt learning.  AST achieves better Last and Average
mAP, CF1, and OF1, as shown in Figure~\ref{fig:figure2}(d).

In summary, DP produces class-conditioned similarities while retaining previous prompts unchanged. AST adapts their tempering strength to the incremental configuration, reducing FPR without modifying learned prompts. Next, we describe the prompting  strategy.

\subsection{Parameter-Efficient Prompting}

DP employs compact class-specific prompts in both visual and textual
modalities. For the text branch, following the common prompt-insertion scheme
~\cite{sun2022dualcoop,hu2023dualcoop++,zhao2024dynamic}, we concatenate the
positive or negative text prompt with the corresponding class name and feed
the resulting sequence into the frozen text encoder. We therefore focus below
on the parameter-efficient implementation of visual prompting. Since
class-conditioned visual encoding introduces more computation, each
visual prompt is injected only into the last $n$ Transformer blocks, as shown in Figure~\ref{fig:figure2}(a) and (b).

\begin{table*}[t]
\centering

\resizebox{\linewidth}{!}{
\begin{tabular}{ll c c c c c c c c c}
\toprule
& & & \multicolumn{4}{c}{\textbf{MS-COCO B40-C10}}
& \multicolumn{4}{c}{\textbf{MS-COCO B0-C10}} \\
\cmidrule(lr){4-7}\cmidrule(lr){8-11}

& \textbf{Method} & \textbf{Type}
& \textbf{Avg.} & \multicolumn{3}{c}{\textbf{Last}}
& \textbf{Avg.} & \multicolumn{3}{c}{\textbf{Last}} \\
\cmidrule(lr){4-4}\cmidrule(lr){5-7}
\cmidrule(lr){8-8}\cmidrule(lr){9-11}

& & & \textbf{mAP} & \textbf{mAP} & \textbf{CF1} & \textbf{OF1}
& \textbf{mAP} & \textbf{mAP} & \textbf{CF1} & \textbf{OF1} \\
\midrule

\multirow{6}{*}{\rotatebox[origin=c]{90}{\textbf{TResNet}}\hspace{6pt}}

& KRT \cite{dong2023knowledge} & M
& 77.8 & 74.0 & 64.4 & 63.4
& 74.6 & 65.9 & 55.6 & 56.5 \\

& AGCN \cite{10221710}  & M
& 73.9 & 69.1 & 58.7 & 59.9
& 72.4 & 61.4 & 53.9 & 56.6 \\

& CSC \cite{du2024confidence} & M
& 78.2 & 75.0 & 65.7 & 67.0
& 78.0 & 72.8 & 64.9 & 66.8 \\

& HCP \cite{zhang2025specifying} & M
& 78.9 & 75.3 & 64.9 & 68.6
& 77.9 & 71.2 & 60.4 & 65.3 \\

& RebLL \cite{du2025rebalancing} & M
& 76.4 & 72.5 & 60.7 & 64.9
& 77.2 & 70.4 & 60.9 & 63.8 \\


& N-CGCN \cite{du2026negative} & M
& 79.2 & 76.3 & 68.1 & \underline{70.4}
& 78.3 & 73.5 & 65.2 & \underline{68.1} \\

\midrule

\multirow{4}{*}{\rotatebox[origin=c]{90}{\textbf{ViT}}\hspace{6pt}}

& L2P \cite{wang2022learning} & S
& 73.7 & 71.1 & 62.8 & 62.3
& 73.4 & 68.0 & 58.8 & 56.0 \\

& DualPrompt \cite{dualprompt2022wang} & S
& 74.3 & 71.4 & 63.0 & 64.3
& 74.1 & 68.6 & 55.3 & 56.3 \\

& CODA-P \cite{codaprompt2023smith} & S
& 73.0 & 73.0 & 65.4 & 66.6
& 75.8 & 70.5 & 59.0 & 59.4 \\

& MULTI-LANE \cite{de2024less} & M
& 78.9 & 76.6 & 65.9 & 66.5
& 79.4 & \underline{74.3} & \underline{65.7} & 64.8 \\

\midrule

\multirow{8}{*}{\rotatebox[origin=c]{90}{\textbf{CLIP}}\hspace{6pt}}

& CL-CLIP \cite{thengane2022clip} & S
& 5.2 & 4.4 & 6.6 & 7.0
& 6.8 & 4.4 & 6.6 & 7.0 \\

& CLAP \cite{jha2024clap4clip} & S
& 55.8 & 56.2 & 62.9 & 62.5
& 65.2 & 55.9 & 62.3 & {66.9} \\

& RAPF \cite{huang2024class} & S
& 70.8 & 64.1 & 58.0 & 59.7
& 69.3 & 59.3 & 54.4 & 54.0 \\

& MOE4CL \cite{yu2024boosting} & S
& 75.4 & 68.4 & 62.2 & 59.7
& 77.9 & 66.9 & 60.0 & 54.0 \\
& DPA \cite{zhao2024dynamic} & M
& \underline{81.1} & \underline{77.2} & \underline{68.2} & 68.5
& \underline{80.7} & 73.3 & 65.1 & 65.1 \\
& MG-CLIP \cite{huang2025mind} & S
& 74.8 & 71.0 & 61.3 & 65.2
& 77.8 & 70.4 & 60.8 & 63.7 \\

& SECA \cite{he2026harnessing} & S
& 73.3 & 69.4 & 57.4 & 59.4
& 71.8 & 70.1 & 59.5 & 61.2 \\

& \textbf{DeCLIP} & M
& \textbf{84.1{\tiny $\pm$0.2}} & \textbf{81.4{\tiny $\pm$0.3}}
& \textbf{70.9{\tiny $\pm$0.5}} & \textbf{71.9{\tiny $\pm$0.4}}
& \textbf{84.0{\tiny $\pm$0.3}} & \textbf{78.3{\tiny $\pm$0.4}}
& \textbf{68.7{\tiny $\pm$0.6}} & \textbf{70.0{\tiny $\pm$0.5}}  \\

\bottomrule
\end{tabular}}
\caption{Comparison with replay-free methods on MS-COCO dataset (\%).
Type S and M denote SLCIL and MLCIL methods, respectively.
Results are grouped by backbone families: TResNet, ViT-B/16, and CLIP with ViT-B/16.
All results of our method are averaged over three runs.
The best results are highlighted in bold, and the second-best results are underlined.}
\label{tab:results_1}
\end{table*}

\noindent\textbf{Visual prompt placement.}
Visual Transformer blocks  transform local patterns into semantic
representations~\cite{raghu2021vision}. We attach visual prompts only to the
last $n$ Transformer blocks, while the first $12-n$ blocks are shared across
all classes and computed once for each image. This design confines
class-dependent computation to high-level visual layers while retaining
class-specific adaptation.

\noindent\textbf{Layer-shared visual-prompt injection.}
Let
$\mathbf{P}_{\mathrm{V}}^{c}\in\mathbb{R}^{L_P\times d}$
denote a class-specific visual prompt, where $L_P$ is the prompt length and
$d$ is the hidden dimension.
Rather than learning a separate visual prompt for each prompted block, we share
the same $\mathbf{P}_{\mathrm{V}}^{c}$ across the last $n$ blocks.

Given the input tokens $\bm{h}\in\mathbb{R}^{L\times d}$ of a multi-head
self-attention (MSA) layer, its output is given by
\begin{equation}
\mathrm{MSA}(\bm{h}_{Q},\bm{h}_{K},\bm{h}_{V})
=
\mathrm{Concat}(\mathrm{h}_1,\ldots,\mathrm{h}_m)\mathbf{W}^{O},
\end{equation}
\begin{equation}
\mathrm{h}_i
=
\mathrm{Attention}\!\left(
\bm{h}_{Q}\mathbf{W}_i^{Q},
\bm{h}_{K}\mathbf{W}_i^{K},
\bm{h}_{V}\mathbf{W}_i^{V}
\right).
\end{equation}
Here, $m$ is the number of attention heads, and $\mathbf{W}^{O}$,
$\mathbf{W}_i^{Q}$, $\mathbf{W}_i^{K}$, and $\mathbf{W}_i^{V}$ are projection
matrices. For self-attention,
$\bm{h}_{Q}=\bm{h}_{K}=\bm{h}_{V}=\bm{h}$. We prepend the class-specific prompt
to the inputs of MSA:
\begin{equation}
\label{eq:prompt_injection}
f(\mathbf{P}_{\mathrm{V}}^{c},\bm{h}) =
\mathrm{MSA}\bigl(
[\mathbf{P}_{\mathrm{V}}^{c};\bm{h}_{Q}],
[\mathbf{P}_{\mathrm{V}}^{c};\bm{h}_{K}],
[\mathbf{P}_{\mathrm{V}}^{c};\bm{h}_{V}]
\bigr).
\end{equation}
The prompt-token outputs are discarded after attention, and only the image
tokens
\begin{equation}
f(\mathbf{P}_{\mathrm{V}}^{c},\bm{h})[L_P:]
\in\mathbb{R}^{L\times d}
\end{equation}
are retained for the subsequent frozen Transformer operations. The same prompt
is reintroduced in each of the last $n$ blocks. By sharing prompts across these
blocks, DeCLIP reduces the number of trainable prompt parameters compared with
learning separate prompts at every prompted layer
~\cite{dualprompt2022wang,codaprompt2023smith}.
Finally, we optimize the prompts with the binary cross-entropy objective:
\begin{equation}
\label{eq:bce}
\mathcal{L}_{\mathrm{BCE}} =
-\sum_{c\in\mc{C}^{t}}
\left[
y_c^t\log(\hat{y}_{c+}^{t})
+
(1-y_c^t)\log(\hat{y}_{c-}^{t})
\right],
\end{equation}
where $y_c^t\in\{0,1\}$ is the ground-truth label of category $c$ at the
current stage, and $\hat{y}_{c+}^{t}$ and $\hat{y}_{c-}^{t}$ are the
probabilities obtained from its positive and negative similarity scores.

\begin{figure}[t]
        \centering
        \includegraphics[width=0.9\linewidth]{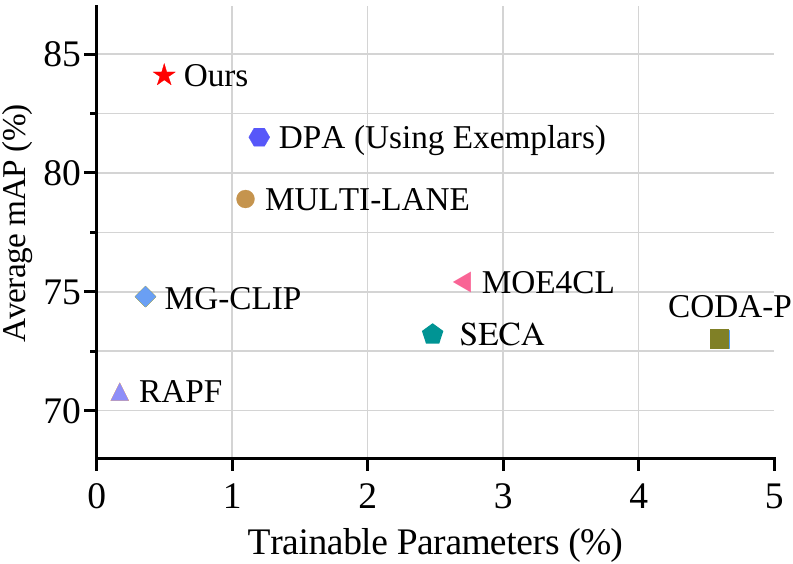}

    \caption{Parameter--performance comparison of methods on MS-COCO B40-C10; percentages use model-specific totals.}
    \label{fig:cost}
    \end{figure}

\section{Experiments}
\subsection{Experimental Settings}
\textbf{Dataset and evaluation.} Following the experimental protocols in  \cite{dong2023knowledge,du2024confidence}, we evaluate our approach on MS-COCO 2014 \cite{lin2014microsoft}, PASCAL VOC 2007 \cite{everingham2010pascal}, and NUS-WIDEseq~\cite{kim2020imbalanced}. MS-COCO contains 82,081 training and 40,504 validation images spanning 80 categories. VOC provides 20 categories with 5,011 training and 4,952 test images. NUS-WIDEseq~\cite{kim2020imbalanced} is a real-world MLCIL benchmark built from NUS-WIDE~\cite{chua2009nus}, containing 49 classes, 48,724 training images, and 2,367 test images. We report mean average precision (mAP), class-wise F1 score (CF1), and overall F1 score (OF1). For each metric, we provide the last model performance (Last) and the average performance (Avg.).

\begin{table*}[t]
\centering

\resizebox{0.95\linewidth}{!}{
\begin{tabular}{ll c c c c c c c c c}
\toprule
& & & \multicolumn{4}{c}{\textbf{VOC B0-C4}}
& \multicolumn{4}{c}{\textbf{VOC B4-C2}} \\
\cmidrule(lr){4-7}\cmidrule(lr){8-11}

& \textbf{Method} & \textbf{Type}
& \textbf{Avg.} & \multicolumn{3}{c}{\textbf{Last}}
& \textbf{Avg.} & \multicolumn{3}{c}{\textbf{Last}} \\
\cmidrule(lr){4-4}\cmidrule(lr){5-7}
\cmidrule(lr){8-8}\cmidrule(lr){9-11}

& & & \textbf{mAP} & \textbf{mAP} & \textbf{CF1} & \textbf{OF1}
& \textbf{mAP} & \textbf{mAP} & \textbf{CF1} & \textbf{OF1} \\
\midrule

\multirow{4}{*}{\rotatebox[origin=c]{90}{\textbf{TResNet}}\hspace{6pt}}

& KRT \cite{dong2023knowledge} & M
& 91.8 & 84.2 & 60.1 & 59.8
& 71.0 & 43.6 & 13.7 & 31.0 \\

& AGCN \cite{10221710}  & M
& 84.3 & 73.4 & 54.6 & 46.8
& 72.1 & 53.4 & 35.3 & 31.7 \\

& CSC \cite{du2024confidence} & M
& 90.4 & 85.1 & 67.7 & 62.2
& 83.3 & 74.1 & 52.4 & 42.9 \\

& RebLL \cite{du2025rebalancing} & M
& 91.4 & 85.1 & 72.6 & 75.8
& 84.6 & 73.1 & \underline{57.2} & \underline{60.0} \\

\midrule

\multirow{4}{*}{\rotatebox[origin=c]{90}{\textbf{ViT}}\hspace{6pt}}

& L2P \cite{wang2022learning} & S
& 89.1 & 82.7 & 55.2 & 51.2
& 85.8 & 77.4 & 54.5 & 53.6 \\

& DualPrompt \cite{dualprompt2022wang} & S
& 89.9 & 83.4 & 60.6 & 58.9
& 86.5 & 78.2 & 43.7 & 44.7 \\

& CODA-P \cite{codaprompt2023smith} & S
& 90.3 & 84.0 & 65.5 & 62.1
& 87.6 & 78.6 & 45.7 & 44.9 \\

& MULTI-LANE \cite{de2024less} & M
& \underline{93.5} & \underline{88.8} & 74.8 & 69.9
& \underline{89.8} & \underline{82.1} & 43.2 & 36.1 \\

\midrule

\multirow{7}{*}{\rotatebox[origin=c]{90}{\textbf{CLIP}}\hspace{6pt}}

& CL-CLIP \cite{thengane2022clip} & S
& 13.8 & 7.7 & 12.4 & 13.2
& 12.9 & 7.7 & 12.4 & 13.2 \\

& CLAP \cite{jha2024clap4clip} & S
& 77.9 & 68.3 & \underline{75.6} & 72.8
& 81.7 & 61.4 & 46.1 & 49.4 \\

& RAPF \cite{huang2024class} & S
& 84.6 & 71.2 & 67.7 & 66.6
& 80.8 & 66.2 & 27.3 & 25.7 \\

&DPA \cite{zhao2024dynamic} & M & \underline{93.5} & 88.1 & - & -
    & 83.5 & 70.9 & 42.8 & 38.0 \\

& MG-CLIP \cite{huang2025mind} & S
& 92.5 & 85.4 & 72.7 & 64.4
& 85.2 & 77.9 & 24.6 & 25.1 \\


& SECA \cite{he2026harnessing} & S
& 92.9 & 88.0 & 75.1 & \underline{77.9}
& \underline{89.8} & 81.6 & 56.0 & 56.5 \\

& \textbf{DeCLIP} & M
& \textbf{95.1{\tiny $\pm$0.2}} & \textbf{90.7{\tiny $\pm$0.3}}
& \textbf{81.7{\tiny $\pm$0.5}} & \textbf{82.7{\tiny $\pm$0.4}}
& \textbf{90.8{\tiny $\pm$0.4}} & \textbf{83.6{\tiny $\pm$0.5}}
& \textbf{71.8{\tiny $\pm$0.5}} & \textbf{74.4{\tiny $\pm$0.6}} \\

\bottomrule
\end{tabular}}
\caption{Comparison with replay-free methods on PASCAL VOC dataset (\%).
}
\label{tab:results_2}
\end{table*}

\begin{figure}[t]
        \centering
        \includegraphics[width=\linewidth]{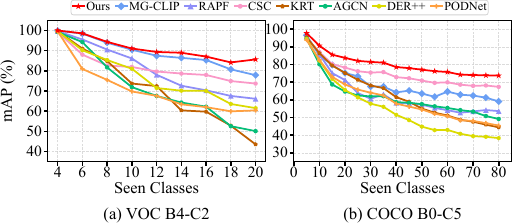}
        \caption{The results of MLCIL on challenging VOC B4-C2 and COCO B0-C5 with more tasks.
}
        \label{fig:figure6}
\end{figure}

\begin{table}[t]
\centering

\resizebox{\linewidth}{!}{
\begin{tabular}{l c c c c c}
\toprule
& & \multicolumn{2}{c}{\textbf{Random}} 
& \multicolumn{2}{c}{\textbf{Sequential}} \\
\cmidrule(lr){3-4}\cmidrule(lr){5-6}

\textbf{Method} & \textbf{\#P}
& \textbf{Avg.} & \textbf{Last}
& \textbf{Avg.} & \textbf{Last} \\

\hline

RebLL
& 29.4M
& 51.7 & 38.4
& 45.2 & 32.8 \\

CSC
& 34.9M
& 55.7 & 44.4
& 50.5 & 40.4 \\

N-CGCN
& 34.9M
& 58.3 & 46.2
& 51.4 & 42.9 \\


MULTI-LANE
& 0.8M
& \underline{59.6} & 52.9
& \underline{63.3} & 53.8 \\

CL-CLIP
& 0.0M
& 7.3 & 5.4
& 7.1 & 5.4 \\

CLAP
& 9.5M
& 34.9 & 23.5
& 36.1 & 25.4 \\

RAPF
& 0.3M
& 55.4 & 48.0
& 59.9 & 49.7 \\

MG-CLIP
& 0.5M
& 58.9 & 49.5
& 61.1 & 50.0 \\


SECA
& 4.3M
& 58.0 & \underline{53.7}
& 57.7 & \underline{54.0} \\

\textbf{DeCLIP}
& {0.5M}
& \textbf{66.8} & \textbf{56.3}
& \textbf{64.3} & \textbf{55.8} \\

\bottomrule
\end{tabular}}
\caption{Comparison of mAP (\%) on the real-world NUS-WIDEseq benchmark under two task orders:
a random order (3, 1, 0, 5, 4, 2) and a sequential order (0, 1, 2, 3, 4, 5).
\#P denotes the number of trainable parameters.}

\label{tab:nuswide_longtail}
\end{table}

\noindent \textbf{Experimental setup.} Following prior MLCIL studies  \cite{dong2023knowledge,du2025rebalancing,zhang2025specifying}, we describe different MLCIL settings using the notation {Bx-Cy}, where “x’’ denotes the number of classes in the base task and “y’’ the number of new classes introduced per incremental task.
 On COCO, we consider two standard scenarios ({B0-C10, B40-C10}). For VOC, we adopt B0-C4 and B4-C2 scenarios. 

\noindent \textbf{Implementation details.}  We use the CLIP ViT-B/16 backbone \cite{radford2021learning}.
For comparison, we include baselines built on the strong TResNet-M \cite{ridnik2021tresnet} (designed for multi-label classification), and methods based on ViT-B/16 \cite{dosovitskiy2020image} and CLIP ViT-B/16.
 The input image size is $224 \times224$. We optimize the prompts for 5 epochs on three datasets using Adam
($\beta_1=0.9$, $\beta_2=0.999$), with a learning rate of $8\times10^{-3}$. The prompt length 
is set to 4, $n=5$ and $\lambda=16$ without test-set tuning. Following \cite{dong2023knowledge,du2024confidence, du2025rebalancing}, classes are ordered lexicographically, and all methods use the same decision threshold. We use the same CLIP logit scale as in~\cite{sun2022dualcoop,zhao2024dynamic}. All DeCLIP results are averaged over three runs.

\subsection{Overall Performance}

We conduct a comparison of replay-free SLCIL and MLCIL using three backbones.
The TResNet-based group consists of KRT \cite{dong2023knowledge}, CSC \cite{du2024confidence}, RebLL \cite{du2025rebalancing}  and SOTA N-CGCN \cite{du2026negative}.
The vision-only group includes  L2P \cite{wang2022learning}, DualPrompt \cite{dualprompt2022wang}, CODA-P \cite{codaprompt2023smith}, and SOTA MULTI-LANE \cite{de2024less}.
The vision–language group comprises CL-CLIP \cite{thengane2022clip}, CLAP \cite{jha2024clap4clip}, RAPF \cite{huang2024class},  MOE4CL \cite{yu2024boosting}, MG-CLIP \cite{huang2025mind}, and DPA \cite{zhao2024dynamic}. The compared results are either taken from prior work~\cite{du2024confidence,zhao2024dynamic,du2026negative} or reproduced using the released implementations.  


\noindent \textbf{MS-COCO.}   
As shown in Table~\ref{tab:results_1}, on MS-COCO, our method consistently surpasses the existing SLCIL and MLCIL methods with three backbones across all evaluation metrics.
On B40-C10, our DeCLIP achieves 84.1\% Avg. mAP and 81.4\% Last mAP, clearly improving over the prior CLIP-based methods (e.g., DPA with 81.1\% Avg. mAP).
Similar gains on B0-C10 show robustness to forgetting as the label space expands.

\noindent \textbf{PASCAL VOC.} Table~\ref{tab:results_2} reports the results on VOC.
 DeCLIP also outperforms previous methods across metrics under both incremental protocols.
On VOC B0-C4, DeCLIP reaches 90.7\% Last mAP, along with 81.7\% CF1 and 82.7\% OF1, exceeding all competitors.
The superiority remains consistent on long-sequence VOC B4-C2, indicating robust knowledge retention over longer task sequences.

\noindent \textbf{Parameter-efficiency analysis.}
As shown in Figure \ref{fig:cost}, we analyze the trade-off between trainable parameters and performance on COCO B40-C10.  Compared with prior CLIP-based approaches such as SECA \cite{he2026harnessing}, MG-CLIP \cite{huang2025mind}, RAPF \cite{huang2024class},  MOE4CL \cite{yu2024boosting}, and DPA \cite{zhao2024dynamic} using 1600 exemplars,
our method achieves the superior average mAP while requiring limited trainable parameters.

\noindent \textbf{Long-sequence performance.} 
Figure \ref{fig:figure6} shows that DeCLIP maintains higher cumulative mAP throughout the long-sequence VOC B4-C2 and COCO B0-C5 settings, supporting reduced forgetting.

\noindent \textbf{Real-world results.}
Since DeCLIP is built on CLIP, evaluating its ability to generalize to
real-world data is important. We therefore conduct experiments on the
long-tailed, web-collected NUS-WIDEseq benchmark. We consider two task schedules: a random task order
$(3,1,0,5,4,2)$ and a sequential order $(0,1,2,3,4,5)$. As shown in
Table~\ref{tab:nuswide_longtail}, DeCLIP consistently achieves better average
and final mAP under both schedules, demonstrating that its CLIP-based
adaptation remains effective in a realistic long-tailed setting.



\begin{table}[t]
\centering

     \resizebox{\linewidth}{!}{
    \begin{tabular}{lccc|ccc}
    \toprule
    \multirow{2}{*}{Method} & \multicolumn{3}{c}{Last} & \multicolumn{3}{c}{Avg.} \\
    \cmidrule(r){2-4}\cmidrule(l){5-7}
     & mAP & CF1 & OF1 & mAP & CF1 & OF1 \\
    \midrule
    CL-CLIP              & 7.7 & 12.4 & 13.2 & 12.9 & 20.1 & 20.6  \\
    \hline
    DP (Text)     & 71.1 & 40.3 & 31.7 & 81.7 & 60.2 & 50.8 \\
    DP (Text + Visual)  & 83.6 & 43.3 & 37.6 & 90.8 & 62.8 & 56.8 \\
    \textbf{DP + AST }& \textbf{83.6} & \textbf{71.8} & \textbf{74.4} & \textbf{90.8} & \textbf{82.2} & \textbf{82.3} \\
\bottomrule
\end{tabular}
}
\caption{
   Ablation of Decoupled Prompting (DP) and Adaptive Similarity
Tempering (AST) on VOC B4-C2 (\%).}

\label{tab:class_specific_disentangle}
\end{table}

\begin{table}[t]
\centering

\resizebox{\linewidth}{!}{
\begin{tabular}{lccc|ccc}
\toprule
\multirow{2}{*}{Configuration} & \multicolumn{3}{c}{Last} &
\multicolumn{3}{c}{Avg.} \\
\cmidrule(r){2-4}\cmidrule(l){5-7}
& mAP & CF1 & OF1 & mAP & CF1 & OF1 \\
\midrule
CL-CLIP & 7.7 & 12.4 & 13.2 & 12.9 & 20.1 & 20.6 \\
\hline
DP (Neg.) & 76.2 & 58.8 & 58.9 & 86.0 & 70.2 & 70.1 \\
DP (Pos.) & 79.7 & 61.8 & 68.2 & 88.5 & 77.4 & 79.4 \\
\textbf{DP (Pos. + Neg.)}
& \textbf{83.6} & \textbf{71.8} & \textbf{74.4} & \textbf{90.8} & \textbf{82.2} & \textbf{82.3 }\\
\bottomrule
\end{tabular}}
\caption{Effect of positive and negative prompt configurations in Decoupled
Prompting (DP) on VOC B4-C2 (\%). AST is applied to all DP configurations.}

\label{tab:pos_neg_prompts}
\end{table}

\begin{table}[t]
\centering

\resizebox{0.9\linewidth}{!}{
\begin{tabular}{lcccccc}
\toprule
& & \multicolumn{2}{c}{\textbf{B40-C10}} & \multicolumn{2}{c}{\textbf{B0-C10}} \\
\cmidrule(r){3-4}\cmidrule(r){5-6}
Method & Memory & Avg. & Last & Avg. & Last \\
\midrule
PODNet  & \multirow{6}{*}{20/class}
& 71.0 & 64.2 & 70.0 & 58.8 \\
DER++  &
& 73.6 & 66.3 & 72.7 & 63.1 \\

CSC&
& 78.7 & 76.7 & 79.6 & 74.8 \\
RebLL &
& 78.3 & 76.0 & 78.6 & 74.0 \\
DPA &
& \underline{81.5} & 78.4 & \underline{81.7} & 76.1 \\
KAR &
& 81.4 & \underline{78.9} & 81.2 & \underline{76.8} \\
\midrule
\textbf{DeCLIP} & 0
& \textbf{84.1} & \textbf{81.4}
& \textbf{84.0} & \textbf{78.3} \\
\bottomrule
\end{tabular}
}
\caption{Comparison of mAP (\%) between our replay-free method and CIL approaches using memory.} 

\label{tab:results_memory_coco}
\end{table}

\subsection{Ablation Study and Analysis}

\textbf{Ablation of different components.} 
Table~\ref{tab:class_specific_disentangle} presents the ablation of
DP and AST. Starting from CL-CLIP, decoupled text
prompting, denoted as DP (Text), already provides a substantial performance
gain. Adding decoupled visual prompting to form DP (Text + Visual) further
improves the results, demonstrating the benefit of learning class-specific
prompts in both modalities. Adding AST reduces the FPR from 25.4\% to 2.4\%.
While mAP remains
unchanged because AST preserves within-class ranking, it substantially improves CF1 and OF1 by suppressing false positives with configuration-dependent tempering strength.
Table~\ref{tab:pos_neg_prompts} further analyzes the positive and negative
prompt design within DP. DP with either negative or positive
prompts already substantially improves over the prompt-free CL-CLIP baseline.
Combining positive and negative prompts achieves the best performance across
all metrics, indicating that jointly modeling class presence and absence
provides more reliable class-wise evidence.

Table~\ref{tab:results_memory_coco} compares our replay-free method
with strong CIL methods using memory, including PODNet~\cite{douillard2020podnet},
DER++~\cite{buzzega2020dark}, CSC,
RebLL, CLIP-based
DPA, and SOTA KAR~\cite{CAO2026133167}.
Despite storing no exemplars, DeCLIP outperforms these approaches.


\begin{figure}[t]
    \centering
    \includegraphics[width=\linewidth]{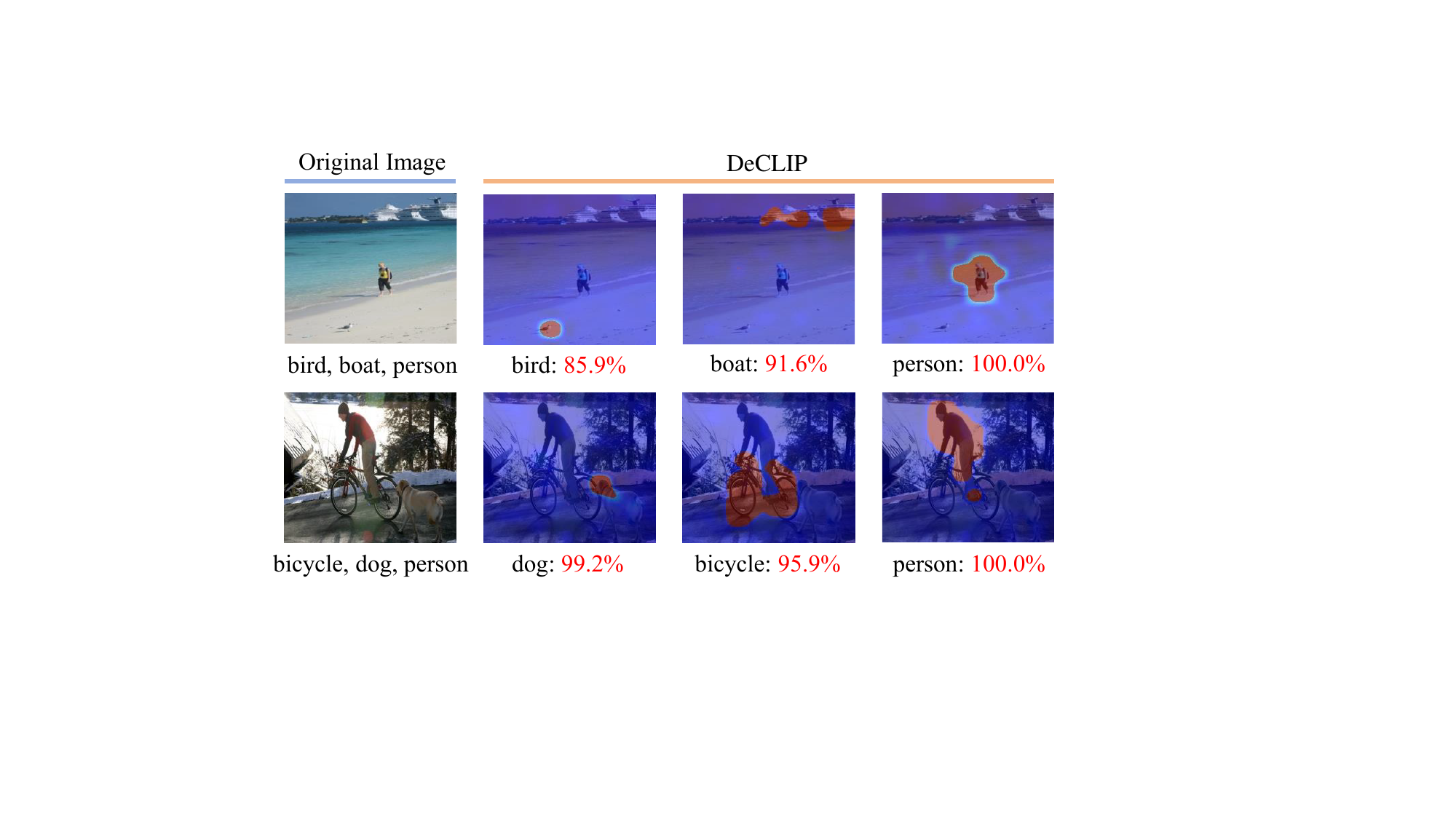}
    \caption{Final-task model visualization.}
    \label{fig:visual}
\end{figure}

\noindent \textbf{Class activation maps.}
  As shown in Figure \ref{fig:visual}, the original images are shown in the first column, followed by class activation maps (CAM) corresponding to each predicted category with confidence scores. For example, in the second row, the model correctly identifies ``bicycle'', ``dog'' and ``person'', and the highlighted regions precisely align with the respective objects. 
  Notably, the body of ``person'', including fine-grained parts such as the legs, can be clearly delineated, demonstrating the  ability to capture detailed  semantics. 



\section{Conclusion}

MLCIL remains challenging for CLIP-based CIL. We find
that applying CLIP to MLCIL can lead to representation entanglement among
co-occurring categories and a high FPR under  partial
labeling. To address these issues, we propose DeCLIP, a replay-free and
parameter-efficient framework for CLIP-based MLCIL. DeCLIP uses DP to learn class-specific prompts in both visual
and textual modalities, producing class-conditioned representations for MLCIL. 
Only new-category prompts are optimized, while previous prompts remain unchanged and are reused at inference, preserving prior knowledge and mitigating catastrophic forgetting without replay.
DeCLIP further incorporates AST, which adaptively adjusts
similarity-tempering strength at inference according to the
incremental configuration. 
To keep the adaptation parameter-efficient, DeCLIP uses compact prompts and
shares each visual prompt across the prompted late layers.
 Extensive experiments on MS-COCO, PASCAL VOC,
and the real-world NUS-WIDEseq benchmark demonstrate that DeCLIP consistently
outperforms prior approaches.

\bibliography{aaai2027}


\end{document}